\documentclass[conference]{IEEEtran}
\usepackage{amssymb}
\usepackage{amsmath}
\usepackage[latin1]{inputenc}
\usepackage{mathabx}
\usepackage{setspace}
\usepackage{psfrag}
\usepackage{color}
\usepackage{cite}
\usepackage{multirow}
\usepackage{wasysym}
\usepackage{graphicx,times,amsmath,balance,amsfonts,cite,pgf,tikz,pgffor}
\RequirePackage{algorithm}
\RequirePackage{algorithmic}
\newtheorem{theorem}{Theorem}

\newtheorem{corollary}[theorem]{Corollary}

\newtheorem{lemma}[theorem]{Lemma}

\usepackage{pgfplots}
\usetikzlibrary{shadows.blur}
\usetikzlibrary{shapes.symbols}

\usepackage{microtype}
\usepackage{graphicx}
\usepackage{subfigure}
\usepackage{booktabs}

\begin{document}
\title{Anytime Stochastic Gradient Descent: A Time to Hear from all the Workers}
\author{\IEEEauthorblockN{Nuwan Ferdinand and Stark C. Draper}
\IEEEauthorblockA{Department of Electrical and Computer  Engineering, University of Toronto, Toronto, ON, Canada}
\IEEEauthorblockA{Email: \{nuwan.ferdinand, stark.draper\}@utoronto.ca }}

\maketitle
\begin{abstract}
In this paper, we focus on approaches to parallelizing
stochastic gradient descent (SGD) wherein data is farmed out to a set of workers, the results of which, after a number of updates, are then combined at a central master node.  Although such \emph{synchronized} SGD approaches parallelize well in idealized computing environments, they often fail to realize their promised computational acceleration in practical settings. One cause is slow workers, termed \emph{stragglers}, who can cause the fusion step at the master node to stall, which greatly slowing convergence. In many straggler mitigation approaches
work completed by these nodes, while only partial, is discarded completely. In this paper, we propose an approach to parallelizing synchronous SGD that exploits the work completed by all workers. The central idea is to fix the computation time of each worker and then to combine distinct contributions of all workers. We provide a convergence analysis and optimize the combination function. Our numerical results demonstrate an improvement of several factors of magnitude in comparison to existing methods.
\end{abstract}

\section{Introduction}

Stochastic gradient descent (SGD) is an optimization algorithm used in many data-intensive machine learning problems \cite{Bertsekas:1997}.
It has recently received significant renewed attention due to the
important role it plays in training deep neural networks
\cite{Keuper:2016N}. However, in such large-scale training problems, it can be infeasible to perform SGD in a single processor due to limited storage and
computation capabilities \cite{Keuper:2016N, Hinton:2012, Dean:2012}. These facts together with the advent of high-performance computing, GPU-accelerators, and computer clusters have driven the development of parallelized variants of SGD \cite{Bertsekas:1997,Martin:NIPS10,Dekel:2012,Hogwild:11,Gemulla:2011}.  Approaches to parallelizing SGD broadly fall into two categories: asynchronous (``Async-SGD'') and synchronous (``Sync-SGD'').

Asynchronous methods are discussed in~\cite{Hogwild:11,Dean:2012,Chilimbi:2014,Remi:2016}. Due to asynchronous memory access, Async-SGD often computes gradient updates from ``stale'' information. Staleness introduces noise (or error) and can lead to poor performance in large-scale problems when one tries to run too many workers in parallel. The focus of this work is on Sync-SGD~\cite{Martin:NIPS10,Dekel:2012,Pan2017,Alex2017:GC}. In Sync-SGD, workers
receive the latest parameter vector in parallel, compute their
gradients, and send their updates to the master node to be
combined. Sync-SGD avoids stale gradients by waiting for all workers
to finish. In comparison to Async-SGD, this reduces the error in the
final output of Sync-SGD. Although Sync-SGD avoids staleness, in
practice, the time to update now depends on the slowest worker. These
slow workers, referred to as \emph{stragglers}, can introduce significant
delays in each combining step, thereby greatly slowing convergence.

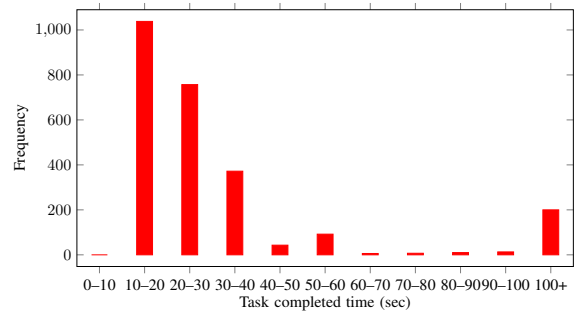
\begin{figure}
	\centering
	\pgfplotstableread[row sep=\\,col sep=&]{
    interval & carT \\
      0--10&	0\\
    10--20&	1038\\
    20--30&	757\\
    30--40&	372\\
    40--50&	43\\
    50--60&	92\\
    60--70&	6\\
    70--80&	7\\
    80--90&	10\\
    90--100&	13\\
    100+&	200\\
    }\mydata

\begin{tikzpicture}[scale=0.6]
\begin{axis}[
ybar,
x=1cm,
enlargelimits=.05,
symbolic x coords={0--10,10--20,20--30,30--40,40--50,50--60,60--70,70--80,80--90,90--100,100+},
xtick=data,
xlabel={Task completed time (sec)},
ylabel={Frequency},
]
\addplot [color=red, fill=red] table[x=interval,y=carT]{\mydata};
\end{axis}
\end{tikzpicture}
	\caption{Histogram of the finishing time of 5000 SGD steps on 20 Amazon EC2 machines.}
	\label{fig:histro}
\end{figure}

We categorize stragglers into two types; \emph{persistent stragglers}
and \emph{non-persistent stragglers}. Persistent stragglers are nodes
that are permanently unavailable or {\em always} take an extremely
long time to complete a task; e.g., due to node failure. On the other
hand, non-persisting stragglers produce an output, but with a randomized
delay in each epoch. Such randomization is often due to shared workloads or heterogeneous networks where distinct physical computers have
differing processing powers. This means that workers finish the same
task in differing amounts of times; often the processing time
distribution has a long tail \cite{Dean:2013}.  To illustrates the effect of stragglers, in Fig.~\ref{fig:histro} we plot a histogram of the finishing times of $5,\!000$ stochastic gradient steps on $20$ Amazon EC2 (Elastic Compute Cloud) nodes. One can observe that the majority of tasks were completed in $10-40$ secs. However, some tasks took more than $100$ secs to complete, resulting in a heavy tail.

In general, stragglers cannot be completely removed from distributed
computing system~\cite{Dean:2013,Lee:TIT18}.  However, there have recently been
a number of approaches that attempt to mitigate the effect of
stragglers in Sync-SGD
\cite{Lee:TIT18,Pan2017,Alex2017:GC,Gauri:AISTAT18,Senur:AX18}. Two of the most
relevant papers are \emph{the fastest $(N-B)$} \cite{Pan2017} and \emph{gradient
coding} \cite{Alex2017:GC}. In \cite{Pan2017}, $N$ workers perform
Sync-SGD. The master node waits for only the first $N-B$ to complete
their tasks before combining. In this way, the finishing time depends
on the fastest $N-B$ out of $N$ workers. It was shown that this
approach can reduce the wall-clock time when the number of stragglers
is fewer than $B$. One drawback to this scheme, as is pointed out in
\cite{Alex2017:GC}, is that in the context of persistent stragglers a
portion of data can be lost. Such loss can introduce a significant error bias
into the final solution (see Fig. 7 in \cite{Alex2017:GC}). In
contrast, in \cite{Alex2017:GC} robustness to persistent stragglers is
introduced by computing each synchronous gradient update at several
worker nodes (say at $S$ workers).  In this way up to $S$ persistent
stragglers can be tolerated. However, the redundant gradient computations are wasteful and consume computational resources. In the presence of only non-persistent stragglers, \cite{Alex2017:GC} shows that such redundant computations can be minimized by coding only part of the data. However, this extension to non-persistent stragglers requires prior knowledge of straggler finishing times and does not work in the combined presence of both persistent and non-persistent stragglers.

In this work, we exploit stragglers rather than avoiding them. Instead of having workers compute a fixed amount of data set in each epoch, we fix the amount of computing time.  The waiting time of the master node is therefore deterministic (up to possible communication delays). The technique is therefore no longer limited by the variability
in finishing times. However, due to the fixed duration of each epoch,
workers complete varying numbers of update steps in each epoch. Thus the
combination step at the master node is non-trivial and needs to incorporate distinct contributions of each worker. We provide a convergence analysis that allows us to optimize the combining factors at the master node. Our method is robust to systems containing both persistent and non-persistent stragglers and automatically adjusts to the realized performance of each
worker. Our proposed scheme effectively introduces data redundancy to
enhance robustness, but at the same time uses that redundancy to accelerate the convergence rather than producing wasteful (unused)
computations.  We perform extensive numerical evaluations by using the
Amazon EC2 and demonstrate the superiority of our method in
comparison to existing techniques. We note that preliminary numerical results of this model is presented in our earlier work \cite{ferdinand:icmla17}. The main contribution of this work is the theoretical convergence analysis, which plays an essential role in practical design. Further, in this paper, we also describe a generalized version of our method that exploits computing resources that previously idled during the period of worker-to-master communication.

\section{Anytime-Gradients}
\label{sec:anytime}
In this section we propose our approach to exploit stragglers in Sync-SGD, which we term \emph{Anytime-Gradients}. The term ``Anytime" is due to the fact that our algorithm can provide a valid solution before all the workers complete their tasks. We first formulate the problem and then describe the algorithm. 
\subsection{Problem formulation}
Our objective is to find an $x \in X$  that minimizes the cost
\begin{align}
\label{eqn:sgd.problem}
F(x) = \sum_{ k= 1}^{m} f_k(x,a_k),
\end{align}
where $f_k(x,a_k)$ is a convex function in $x \in X$ for every data sample $a_k \in A$ and $X$ is a closed convex set. $A$ is the data set with cardinality $|A|=m$. Linear and logistic regression are well known problems that take this form. E.g., for linear regression $f_k(x,a_k)=({b}_k^Tx-y_k)^2$ where $a_k=(b_k,y_k)$. 

The system we consider consists of a master and $N$ worker nodes. The Anytime-Gradients algorithm is detailed in Algorithms 1 and 2. We now discuss the key steps.
\subsection{Data partition and allocation}
We decompose the data set $A$ into $N$ data blocks, denoted by $A_i$, $i \in [N]$ where $[N]=\{1,\ldots N\}$. Note that $|A_i|=m/N$. We assume that the number of persistent stragglers is less than or equal to $S$; a design parameter that determines the robustness of our scheme to persistent stragglers. At initialization, $S+1$ data blocks are
distributed to each worker in a manner such that each block is
distributed to $S+1$ workers. A satisfying assignment can easily be
obtained by circularly shifting data blocks among workers, as is
indicated in Table~\ref{tbl:data.assigment}.  In the Table~\ref{tbl:data.assigment}, $W_i$ is
the $i$-th worker, and x (or o) denotes whether (or not) a particular data block
is assigned to the corresponding worker. We note that one can make different data assignments from that described by Table \ref{tbl:data.assigment}. The important aspect of any data allocation scheme is that each data block is provided to $S+1$ workers.
\begin{table}
	\centering
	\caption{Data assignment to workers}
	\label{tbl:data.assigment}
	\begin{tabular}{|l|l|l|l|l|l|l|l|l|}
		\hline
		& $A_1$ & $A_2$ & $\ldots$ & $A_{S+1}$ & $A_{S+2}$ & $\ldots$ & $A_{N}$ \\ \hline
		$W_1$   & x & x    & x    & x     &   o     &  o    & o    \\ \hline
		$W_2$   & o  & x   &  x  & x    & x    &  o      &  o        \\ \hline
		$\vdots$   &    &   &    &      &      &        &          \\ \hline
		$W_N$   & x & x  & x   &   o   &  o    &     o   & x \\ \hline
	\end{tabular}
\end{table}
Note that in Algorithm \ref{Alg:worker}, $\bar{A}_v$ $v \in [N]$ denotes the portion of data that $v$-th worker
received based on the assignments defined in Table
\ref{tbl:data.assigment}. E.g., for the first worker $\bar{A}_1 = (A_1,A_2, \ldots A_{S+1})$.
\begin{algorithm}
	\caption{$\mathrm{MasterNode}(A, x_0, \eta, \tau, N, S, T, T_c )$}
	\label{Alg:master}
	\begin{algorithmic}[1]
		\STATE \textbf{Input:} Data set $A$, initial parameter vector $x_{0}$, step size $\eta$, number of epochs $\tau$, number of workers $N$, amount of redundancy $S$, predefined computation time $T$, waiting time $T_c$.
		\STATE The master node decomposes data set $A$ into $N$ equally sized data blocks $A_1, \ldots A_N$.
		\FORALL{ $v=1,2,\ldots N$ }
		\STATE  sends data blocks to $v$-th worker based on Table \ref{tbl:data.assigment}.  
		\ENDFOR
		\FORALL {$t=1,\ldots, \tau$}
		\STATE Call workers in parallel: $\mathrm{WorkerSGD} (x_{t-1},\eta, T)$
		\WHILE {\emph{waiting time} $\leq T_c$}
		\STATE Receive $(x_{vt}, q_v)$ from workers, $v=1,2,\ldots N$
		\ENDWHILE
		\STATE Let $\chi$ be the set of workers whose updates were received 
		\IF {$v \notin \chi$ }
		\STATE $x_{vt}=0$, $q_v=0$ and $\lambda_v=0$
		\ENDIF
		\STATE Combine updates $x_{t}= \sum_{v=1}^N \lambda_v x_{vt}$.
		\ENDFOR
		\STATE \textbf{Return:} $x_{\tau}$
	\end{algorithmic}
\end{algorithm}
\begin{algorithm}
	\caption{$\mathrm{WorkerSGD}(x,\eta_v,  T )$}
	\label{Alg:worker}
	\begin{algorithmic}[1]
		\STATE \textbf{Input:} Data $\bar{A}_v$, parameter vector $x$, step size $\eta_v$, and predefined time $T$.
		\STATE Start a runtime counter $\bar{T}$.
		\STATE Set a counter $t=0$.
		\STATE Set $x_{v0}=x$.
		\WHILE {$t\leq m(S+1)/N \;\text{or}\; \bar{T}\leq T$}
		\STATE sample a data point randomly (uniform) from $\{1,2,\ldots m(S+1)/N\}$.
		\STATE update: $x_{vt}  = x_{v(t-1)}-\eta_{vt}\nabla f_{vt}(x_{v(t-1)},a_{vt})$
		\STATE $t=t+1$.
		\ENDWHILE
		\STATE \textbf{Return:} $(x_{vt}, t)$
	\end{algorithmic}
\end{algorithm}
\subsection{Computation and Waiting times}
The computation time $T$ and waiting times $T_c$ are two important parameters of our scheme. $T$ sets the computing duration of each epoch. $T_c$ is set to avoid excessive delays due to worker nodes that suffer from long communication times or that fail completely.   

\subsection{Combining operation at the master node}
One key element of Anytime-Gradients is the combining operation at the
master node.  The combining factors are denoted by $\lambda_v$, $v \in [N]$ in step 15 of Algorithm \ref{Alg:master}. Our objective is to choose these factors that maximize the rate of convergence. For an example, in a situation where
all workers finish the same number of gradient steps, it makes sense to set
$\lambda_v=1/N$ for all $v$. This is uniform averaging and is used in classical Sync-SGD \cite{Martin:NIPS10}. However, this may not be the best selection when workers compute different number of gradient steps, as will
typically be the case in our scheme.

In our setting, workers that computed a large number of gradient steps get closer to the optimal solution than those that complete fewer steps. We therefore differentially weight the outputs of the workers
when combining at the master node. It is a-priori unclear how to find the optimal choice of the $\lambda_v$. Based on several theoretical assumptions, we can find a solution to $\lambda_v$. One such a solution is proved in Theorem \ref{thm:optimal.parameters} and it is given by  
\begin{align}
\lambda_v = \frac{q_v}{\sum_{v=1}^N q_v}, \;\;\;  \forall v \in [N]. \label{eq.weightFactor}
\end{align}

\begin{figure}%
	\centering
	\subfigure[]{\label{fig:different_jobs} \begin{tikzpicture}[scale=.4]
\begin{axis}[
ylabel=\footnotesize{Number of iterations per epoch},
xlabel=\footnotesize{Worker index},
enlargelimits=0.05,
ybar,
]
\addplot [color=red, fill=red]
coordinates {(1,10000) 
			(2,8500)
			(3,8000)
			(4,7500)
		    (5,7250)
	    	(6,6800)
    		(7,5500)
    		(8,2000)
    		(9,1500)
    		(10,500)};
\end{axis}
\end{tikzpicture}}
	\subfigure[]{\label{fig:scale-performance}\tikzset{every mark/.append style={scale=1.5}}
\begin{tikzpicture}[scale=0.4]
	\begin{semilogyaxis}[
		height=8cm,
		width=10cm,
		grid=major,
		xlabel=Epochs,
		ylabel=Normalized error,
		axis on top,xmin=1, xmax=10, ymin=0, ymax=1]
		\addlegendentry{$\lambda_v=1/N$}
		\addplot [line width=0.5mm, color=blue, solid, every mark/.append style={solid, fill=blue},mark=diamond*] coordinates {
		(1,	1.16364)
		(2,	0.965629)
		(3,	0.802209)
		(4,	0.667692)
		(5,	0.556285)
		(6,	0.464065)
		(7,	0.386752)
		(8,	0.322571)
		(9,	0.269343)
		(10,	0.224677)
		};
		\addlegendentry{$\lambda_v$ based on Theorem~\ref{thm:optimal.parameters}}
		\addplot [line width=0.5mm, color=magenta, dashed, every mark/.append style={solid, fill=magenta},mark=asterisk] coordinates {
		(1,	1.16313)
		(2,	0.910305)
		(3,	0.714002)
		(4,	0.561175)
		(5,	0.441976)
		(6,	0.348339)
		(7,	0.274878)
		(8,	0.217595)
		(9,	0.172103)
		(10,	0.136316)
		
		};
	\end{semilogyaxis}
\end{tikzpicture}}
	\caption{(a) Number of iterations performed in an epoch at each of $10$ workers. (b) Normalized error vs epochs for different scaling factors.}
\end{figure}
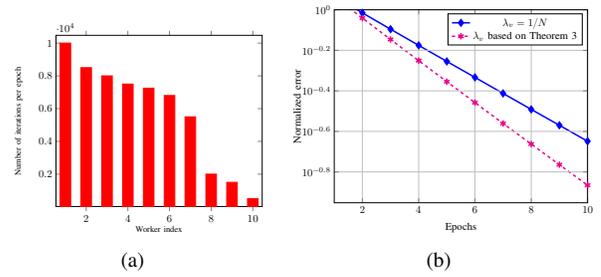

The choice in \eqref{eq.weightFactor} sets each weight proportional to the amount of work completed by the respective worker. In order to test this choice numerically, we perform linear regression using $10^5$ synthetic data samples on $10$ parallel workers. The elements of the data matrix $A\in \mathbb R^{10^5\times 10^3}$ and true parameter vector $x \in \mathbb R^{10^3}$ were generated according to an independent and identically distributed (i.i.d.) Gaussian distribution with zero mean and unit variance, $\mathcal N(0,1)$. The label vector is $ y = Ax + z$ where $z$ is i.i.d.~Gaussian noise generated according to the $\mathcal N(0,10^{-3})$ distribution. We allocate $10^4$ data vectors to each worker and make workers to process different numbers of iterations as shown in Fig. \ref{fig:different_jobs}. In this experiment, in one epoch $W_1$ got through $10,\!000$ data samples, $W_2$ worked through $8500$, whereas the last worker only got through $500$. The error performance is shown in	Fig.~\ref{fig:scale-performance} using the both the choice of $\lambda_v$ from Theorem~\ref{thm:optimal.parameters} and uniform averaging, i.e., $\lambda_v=1/N$. It is evident that using a proportional weighting based on the amount of work completed, i.e., \eqref{eq.weightFactor} leads to far faster converge than simple averaging.

\subsection{Comparison to existing methods} 
We now discuss key conceptual advantages of our proposed
Anytime-Gradients, when compared to two alternative strategies: the fastest $(N-B)$ (FNB) \cite{Pan2017} and Gradient Coding \cite{Alex2017:GC}.

One advantage of the FNB approach is that the finishing time depends
only on the first $N-B$ out of $N$ workers. Hence, up to $B$ stragglers can be avoided. In our scheme, we can achieve the same finishing time by
properly fixing the pre-defined time ($T$), e.g., to match the
$(N-B)$-th order statistic. In this way, we not only expect that
$N-B$ workers will finish all their updates, but the master node also gets to use the
(smaller number of) updates completed by the $B$ slowest workers. Our
method therefore yields faster convergence than FNB.
Moreover, the replication in data placement performed by
Anytime-Gradients makes it robust to persistent stragglers. Therefore Anytime-Gradients does not suffer the abrupt degradation in performance due to data lost when
compared to FNB.

Gradient Coding~\cite{Alex2017:GC} introduces redundant gradient computations to avoid persistent
stragglers. Many of these redundant computations are wasteful in that they do not contribute to the final output. On the other hand, while our method also
introduces redundancy, it does so in a manner such that all redundant computations contribute to faster convergence, without decreasing robustness. Vanilla gradient coding works rather poorly in the presence of non-persistent stragglers and so a second scheme is proposed in~\cite{Alex2017:GC} to handle non-persistent stragglers. This latter scheme requires prior information of the performance of worker nodes (e.g., that stragglers are a certain fixed factor slower than non-stragglers) and does not provide robustness to persistent stragglers. Our proposed scheme, on the other
hand, does not need such prior information and
effectively handles both persistent and non-persistent stragglers.

\section{Convergence analysis}
In this section, we provide a convergence analysis of our scheme. We first find the expected distance to the optimal solution, then we study the variance. Next we find combining factors that minimize the variance to the optimal distance. Finally, we provide high probability bound for the expected distance. In this analysis, we assume that all workers sample from the entire data set. This assumption is important as it allows us to consider a single distribution that convergence is taken with respect to.  In practice, we cannot sample uniformly and data replication, per the replicate $S$ parameter, allows us to approximate this (and importantly to avoid the situation that the only copy of a portion of data is available at what turns out to be a straggler node). We note that the assumption we make here is not an uncommon one.
\subsection{Preliminaries}

As mentioned earlier, we assume $X$ is a convex set and $f_k(x,a_k)$ is convex and differentiable in $x \in W$ for all $k \in [m]$. In much of the following we drop the subscript $k$ for notation convenience. Let $\nabla f(x,a)$ be the gradient of $f(x,a)$ with respect to $x$. We assume the gradient of $f(x,a)$ is Lipschitz continuous, i.e.,
\begin{align}
\|\nabla f(x,a)-\nabla f(\tilde x,a)\| \leq L\|x-\tilde x\|, \;\;\; \forall a, \forall x,\tilde x \in X,
\end{align}
where $\|\cdot \|$ denote $l_2$ norm. We assume that
\begin{align}
\label{eqn:unbiased}
F(x)=E[f(x,a)]
\end{align}
and $\nabla F(x)=E[\nabla f(x,a)]$. In these cases, the expectation is with respect to $a$ where $a$ is uniformly distributed across all $|A|=m$ data samples. Unless otherwise stated expectation will remain with respect to $a$ throughout. We assume there exists a constant $\sigma$ such that $E[\| \nabla f(x,a) - \nabla F(x) \|^2] \leq \sigma^2$. We define $d(x,u)= \frac{1}{2}\|x-u\|^2$ and $D^2 = \max_{x,u \in X} d(x,u)$. We further define the global optimum, $x^*=\arg \min_{x \in X} F(x)$.

\subsection{Expected distance}

The proof is based on stochastic approximation. Suppose the $v$-th worker samples a sequence of data $a_{v0}, a_{v1}, a_{v2}, \ldots $ drawn uniformly at random from entire data set $A$. By observing $a_{v(t-1)}$, the $v$-th worker predicts $x_{vt} \in X$. At the end of each epoch, the output of the $v$-th worker is $x_v = \frac{1}{q_v}\sum_{t=0}^{q_v} x_{vt}$
where $t$ denotes the (sub-epoch) index of iteration. This $x_v$ is a stochastic approximation to the output of Algorithm \eqref{Alg:worker}. The combined parameter vector at the master node is
\begin{align}
x = \sum_{v=1}^{N} \lambda_v x_v = \sum_{v=1}^{N}  \frac{\lambda_v}{q_v}\sum_{t=0}^{q_v} x_{vt}.
\end{align}  
We want to characterize how close this combined parameter vector is to the global minimizer as a function of the choice of $\lambda_v$.

{\theorem \label{thm:expected.distance} Let $x_0$ be the initialized parameter vector provided to all workers and let the step size of the $v$-th worker at the $t$-th iteration  be $\eta_{vt} = L+\sqrt{t+1}\sigma/D$. Then,
	\begin{align}
	& E[F(x) -F(x^*)]   \leq \nonumber \\ & \sum_{v=1}^{N} \frac{\lambda_v}{q_v} \left \{F(x_0) - F(x^*)+LD^2 +  2 \sigma D \sqrt{q_v}  \right\}.
	\end{align} }
\emph{Proof:} See Appendix \ref{apn.expected.distance}. One possible approach is to find combining factors that minimize the expected distance. As the fastest worker has the lowest expected distance, this results in picking the fastest worker only. Therefore, this is not the right approach. The expected distance can in fact be misleading as the it may have a larger variance. Next we study the variance.
\subsection{Variance of the distance to the optimum}
\label{sec:varince}
In this section, we bound the variance to the optimum from the combined solution that is available to the master node. This will be useful in optimizing the combining factors the master node should use. 
{\theorem \label{thm.variance} Let $\mathcal V[\cdot]$ denotes the variance of the argument. Further, assume that $\|\nabla f(x,a)\|\leq G$. Then,
	\begin{align}
	\label{eqn:Tvar3}
	\mathcal V[F(x) -F(x^*)] \leq 2\sigma^2D^2  \left(\frac{  G^2 }{\sigma^2} + 2\right) \sum_{v=1}^{N} \frac{\lambda_v^2}{q_v}.
	\end{align}}
\emph{Proof:} See Appendix \ref{apn:variance}. 

In \eqref{eqn:Tvar3}, the terms before the summation are constant and $q_v$ is the number of gradient steps by the $v$-th worker. We find the combining factors $\lambda_v$ that minimize the variance of the distance to the global minimum $\mathcal V[F(x) -F(x^*)]$.

{\theorem \label{thm:optimal.parameters}
	Let $x^* = \min_{x \in X} F(x)$ and $x = \sum_{v=1}^{N}\lambda_v x_v$ be the combined parameter vector at the master node in a given epoch. Then, the following choice of $\lambda_v$ minimizes the variance of $F(x)-F(x^*)$:
	\begin{align}
	\lambda_v = \frac{q_v}{\sum_{v=1}^N q_v}, \;\;\;  \forall v \in [N]. \label{eq.weightFactor.0}
	\end{align}}
{\emph{Proof:} We write \eqref{eqn:Tvar3} in vector form as
\begin{align}
\label{eqn:var5}
\mathcal V[F(x) -F(x^*)] \leq \frac{1}{2} \lambda^T R \lambda
\end{align}
where $\lambda=[\lambda_1,\ldots \lambda_N]^T$ and $ R$ is a $N\times N$ diagonal matrix with $v$-th element  $r_{vv}= \frac{4\sigma^2D^2}{q_v}  \left[\frac{  G^2 }{\sigma^2} + 2\right]$. 
We optimize the choice of $\lambda$ by minimizing the upper bound \eqref{eqn:var5} on $\mathcal V[F(x) -F(x^*)]$:
\begin{equation*}
\begin{aligned}
& \underset{\lambda_1,\ldots \lambda_N}{\text{min}}
& &  \frac{1}{2} \lambda^T  R \lambda  \\
& \text{subject to}
& & \mathbf 1 \lambda=1 \\
& & & \lambda_v \geq 1, \;\; \forall v \in [N].
\end{aligned}
\end{equation*}
This is quadratic programing with equality constraint and $R$ is a positive semidefinite matrix. 
As $R$ is diagonal matrix, the solution can easily be found and is given in \eqref{eq.weightFactor.0}. 
{\corollary \label{cor:variance}Let $\lambda_v$ is chosen according to the Theorem \ref{thm:optimal.parameters} and $Q=\sum_{v=1}^{N}q_v$. Then 
\begin{align}
\label{eqn:Cvar3}
\mathcal V[F(x) -F(x^*)] \leq \frac{2\sigma^2D^2  \left(\frac{  G^2 }{\sigma^2} + 2\right)}{Q}
\end{align}} One can notice from Corollary \ref{cor:variance} that the variance decays inversely proportional to the total iterations taken by the workers.   
\subsection{High-probability bound}
In the previous section, we derived a bound on the expected distance and its variance. In the following theorem provides a high probability bound on $F(x) -F(x^*)-E[F(x) -F(x^*)]$, i.e., the difference between true distance to expected distance. 
{ \theorem \label{thm.high.probability} Let $\gamma=\max_v{\frac{\lambda_v}{q_v}}$. For any $\delta \in (0,1]$, the $F(x) -F(x^*)-E[F(x) -F(x^*)]$ is bounded with probability at least $1-\delta$ by
	\begin{align}
	&F(x) \!-\!F(x^*) \!- \!E[F(x) \!-\!F(x^*)] \nonumber \leq  \gamma 2GD \left(\frac{G}{\sigma}+2\right) \nonumber \\ & \log(1/\delta) \sqrt{1+\frac{36\sum_{v=1}^{N} \frac{\lambda_v^2\sigma^2D^2}{q_v}  \left(\frac{  G^2 }{\sigma^2} + 2\right)}{\log(1/\delta)}}.
	\end{align}
}
{\corollary Let $Q=\sum_{v=1}^{N}q_v$. Then, the solution for $\lambda_v$ from Theorem \ref{thm:optimal.parameters} yields
	\begin{align}
	\label{eqn:col6}
	&F(x) -F(x^*) - E[F(x) -F(x^*)]\nonumber  \leq  \frac{2GD}{Q} \left(\frac{G}{\sigma}+2\right) \nonumber \\ & \log(1/\delta) \sqrt{1+\frac{36\sigma^2D^2  \left(\frac{  G^2 }{\sigma^2} + 2\right)}{Q \log(1/\delta)}}.
	\end{align}}
\emph{Proof:} The proof is based on Bernstein-type inequality for martingales \cite{Cesa-Bianchi:2006} and provided in Appendix \ref{app.high.probability}. Note that \eqref{eqn:col6} shows that the uncertainty of the expected distance decays roughly inversely proportional with the number $Q$ of total iterations by all workers.

\section{Numerical Results}
\label{sec:numerical}
In this section, we numerically evaluate Anytime-Gradients 
on the Amazon EC2 cloud. We compare results with existing schemes
\cite{Pan2017,Alex2017:GC}. In order to make a fair comparison across simulations, we
conducted all experiments in parallel (i.e., at the same time). This minimizes fluctuation in uncontrollable experimental conditions.
Most simulation results are for linear regression based on
synthetic data. In addition, we test our scheme using real data (a subset of the ``Million Song Dataset" \cite{Lichman:2013}). The synthetic data ($A$) and true
parameter vector ($x^*$) are generated in an i.i.d. manner according to the $\mathcal
N (0,1)$ distribution. The label data used in linear regression is $y=A x +z$ where $z$ is i.i.d.~Gaussian noise $\mathcal N(0,10^{-3})$. The normalized error computed at the
end of the $t$-th epoch is $\|Ax_t- Ax^*\|/\|Ax^*\|$. In all our experiments, we let the waiting time $T_c$ be long enough to receive updates from all workers.

The first experiment we conduct is a linear regression problem where the data
matrix $A$ is $500,\!000\times 1000$. We use $10$ worker nodes
and allocate $50,\!000$ data points to each worker. We do not introduce redundancy, i.e., $S=0$. We fix $T=200$ secs. We
compare results with classical Sync-SGD where the master node
waits for all the workers to finish before
combining. Fig.~\ref{fig:err.wall.500k.10w} plots the error vs.~wall
clock time. It is evident that the Anytime-Gradients
reaches the optimal solution $300$ secs faster than ``wait-for-all'' classical Sync-SGD. 
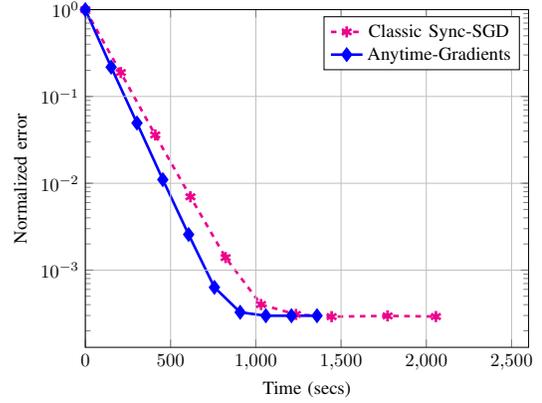
\begin{figure}
	\centering
	\tikzset{every mark/.append style={scale=1.5}}
\begin{tikzpicture}[scale=0.7]
	\begin{semilogyaxis}[
		height=8cm,
		width=10cm,
		grid=major,
		xlabel=Time (secs),
		ylabel=Normalized error,
		axis on top,xmin=0, xmax=2600, ymin=0, ymax=1]
		\addlegendentry{Classic Sync-SGD}
		\addplot [line width=0.5mm, color=magenta, dashed, every mark/.append style={solid, fill=magenta},mark=asterisk] coordinates {
			(0,	1)
			(207.993,	0.188474)
			(409.917,	0.036038)
			(616.299,	0.00698833)
			(822.706,	0.00139341)
			(1030.733,	0.000401409)
			(1235.272,	0.000308741)
			(1444.522,	0.000291102)
			(1772.79,	0.000296954)
			(2056.216,	0.000291834)
		};
	
		\addlegendentry{Anytime-Gradients}
		\addplot [line width=0.5mm, color=blue, solid, every mark/.append style={solid, fill=blue},mark=diamond*] coordinates {
		(0,	1)
		(151.46,	0.21758)
		(302.942,	0.0494003)
		(454.372,	0.0110173)
		(605.839,	0.00257022)
		(757.188,	0.0006327)
		(908.159,	0.000326861)
		(1058.855,	0.000298145)
		(1208.896,	0.000298145)
		(1358.923,	0.000298145)
		
		};

	\end{semilogyaxis}
\end{tikzpicture}
	\caption{Error vs wall-clock time for $5\times 10^5$ data
		points for $10$ workers.}
	\label{fig:err.wall.500k.10w}
\end{figure}

In the next experiment, we introduce redundancy of two, i.e.,
$S=2$. Each worker now gets a $16,\!666 \times 1000$ unique data matrix and an additional $33,\!000\times 1000$ redundant data matrix. In the Anytime-Gradients scheme, we give each worker $T=100$ secs to work in
each epoch. We compare the error performance of Anytime-Gradients with the FNB ($N=10$ and $B=8$)~\cite{Pan2017} and Gradient Coding \cite{Alex2017:GC} in Fig. \ref{fig:fig14}. A significant
improvement is observed in the performance of the proposed Anytime-Gradients. E.g. an error rate of $10^{-0.4}$ is obtained by our scheme in about $100$ and $600$ fewer seconds when respectively compared to FNB~\cite{Pan2017} and Gradient Coding \cite{Alex2017:GC}.
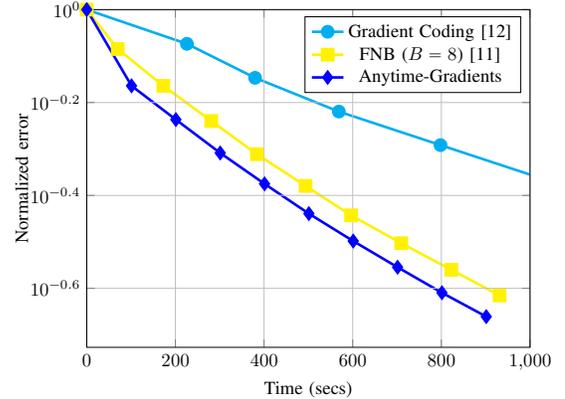
\begin{figure}
	\centering
	\tikzset{every mark/.append style={scale=1.5}}
\begin{tikzpicture}[scale=0.7]
	\begin{semilogyaxis}[
		height=8cm,
		width=10cm,
		grid=major,
		xlabel=Time (secs),
		ylabel=Normalized error,
		axis on top,xmin=0, xmax=1000, ymin=0, ymax=1]
		\addlegendentry{Gradient Coding \cite{Alex2017:GC}}
		\addplot [line width=0.5mm, color=cyan, every mark/.append style={solid, fill=cyan},mark=otimes*] coordinates {
			(0,	1)
			(225.9704,	0.844078)
			(379.7591,	0.713231)
			(568.7147,	0.603309)
			(798.0739,	0.510874)
			(1026.3303,	0.433058)
			(1249.9471,	0.367486)
			(1463.7896,	0.312164)
			(1690.7149,	0.265448)
			(1911.3873,	0.225948)
		};
		\addlegendentry{FNB ($B=8$) \cite{Pan2017}}
		\addplot [line width=0.5mm, color=yellow, solid, every mark/.append style={solid, fill=yellow},mark=square*] coordinates {
			(0,	1)
			(70.0003,	0.822586)
			(172.7563,	0.685407)
			(280.9393,	0.575767)
			(383.8973,	0.488359)
			(493.6093,	0.417398)
			(596.4493,	0.360698)
			(709.4913,	0.314107)
			(822.3773,	0.275329)
			(931.2793,	0.242498)
			
		};
		\addlegendentry{Anytime-Gradients}
		\addplot [line width=0.5mm, color=blue, solid, every mark/.append style={solid, fill=blue},mark=diamond*] coordinates {
		(0	,1)
		(101.039	,0.685496)
		(201.058,	0.579757)
		(301.109,	0.491338)
		(401.14	,0.421606)
		(501.18	,0.363764)
		(601.212,	0.317488)
		(701.257,	0.278853)
		(801.293,	0.24572)
		(901.326,	0.218285)
		};

	\end{semilogyaxis}
\end{tikzpicture}
	\caption{Normalized error vs wall-clock time for $10$ workers
		with redundant data. Each data block is repeated $3$ times
		such that $S=2$. Each worker is given $T=100$ secs. Each
		data block sized $16,666\times 1000$.}
	\label{fig:fig14}
\end{figure}

We test Anytime-Gradientss using real data. We use the ``YearPredictionMSD" data set \cite{Lichman:2013} that predicts the release years of songs via linear regression. The data matrix is $515,\!345\times 90$ and is divided among $10$ parallel workers. We assume $S=1$ so that each data block is repeated at two workers. For Anytime-Gradients we allow each worker $T=20$ secs to update its parameter vector in each epoch. In Fig. \ref{fig:real} we compare normalized error vs.~wall-clock time with FNB ($B=8$) scheme and classical Sync-SGD. It is observed that Anytime-Gradients outperforms the other schemes.       
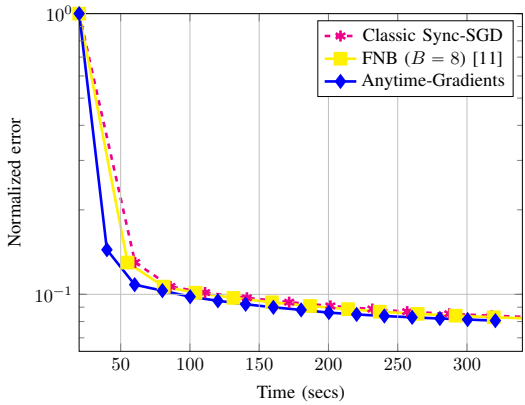
\begin{figure}
	\centering
	\tikzset{every mark/.append style={scale=1.5}}
\begin{tikzpicture}[scale=0.7]
	\begin{semilogyaxis}[
		height=8cm,
		width=10cm,
		grid=major,
		xlabel=Time (secs),
		ylabel=Normalized error,
		axis on top,xmin=20, xmax=340, ymin=0, ymax=1]
		\addlegendentry{Classic Sync-SGD}
		\addplot [line width=0.5mm, color=magenta, dashed, every mark/.append style={solid, fill=magenta},mark=asterisk] coordinates {
			(20.038,	1)
			(50.418+10,	0.129917)
			(70.9029+15,	0.106664)
			(90.9291+20,	0.101424)
			(116.1163+25,	0.0972021)
			(141.644+30,	0.0937568)
			(166.4071+35,	0.0909361)
			(191.7342+40,	0.0886354)
			(211.8332+45,	0.0867655)
			(237.2183+50,	0.0852313)
			(262.7885+55,	0.0839522)
			(287.9171+60,	0.0828827)
			(313.3452+65,	0.0819908)
			(338.4175+70,	0.0812353)
		};
		\addlegendentry{FNB ($B=8$) \cite{Pan2017}}
		\addplot [line width=0.5mm, color=yellow, solid, every mark/.append style={solid, fill=yellow},mark=square*] coordinates {
			(20.038,	1)
			(44.9935+10,	0.12976)
			(66.0382+15,	0.106523)
			(84.0282+20,	0.101323)
			(106.4292+25,	0.0971234)
			(129.1758+30,	0.0936981)
			(151.6684+35,	0.090881)
			(174.1027+40,	0.0885917)
			(191.9942+45,	0.0867294)
			(214.3566+50,	0.0851887)
			(236.8172+55,	0.0839229)
			(259.3226+60,	0.0828641)
			(281.7984+65,	0.0819806)
			(304.0607+70,	0.0812199)
		};
		\addlegendentry{Anytime-Gradients}
		\addplot [line width=0.5mm, color=blue, solid, every mark/.append style={solid, fill=blue},mark=diamond*] coordinates {
		(20.038,	1)
		(40.0736,	0.144196)
		(60.1051,	0.108242)
		(80.1294,	0.10306)
		(100.158,	0.098078)
		(120.1928,	0.094852)
		(140.2267,	0.0921604)
		(160.2617,	0.0899055)
		(180.2845,	0.0880172)
		(200.3157,	0.0860766)
		(220.3428,	0.0847978)
		(240.376,	0.0837115)
		(260.4124,	0.0827921)
		(280.4487,	0.0820054)
		(300.4717,	0.0813212)
		(320.5007,	0.0805946)
		};

	\end{semilogyaxis}
\end{tikzpicture}
	\caption{Normalized error vs wall-clock time for $10$ workers
		for real data. Each data block is repeated $2$ times
		such that $S=1$. Each worker is given $T=20$ secs. Each
		data block sized $515345\times 90$.}
	\label{fig:real}
\end{figure}

\section{Generalized Anytime-Gradients}
\label{sec:generalized}
We now extend Anytime-Gradients to exploit computing
resources that, in our original scheme, idle during the inter-epoch periods of communication
between workers and master. In this manner we can improve the rate
of convergence, though worker nodes will no longer be synchronized at the beginning of each epoch.

Generalized Anytime-Gradients works as follows. As in Anytime-Gradients, workers work for a fixed time $T$ in each epoch, sending updates $x_{vt}$ to the master in parallel. The parameters $v$ and $t$ respectively denote worker and ``epoch" indexes. In Anytime-Gradients, workers then remain idle until they receive the combined updated parameter vector $x^t$ from the master.  In contrast, in the generalized version, workers continue to update $x_{vt}$ during the idle period until they receive the combined parameter vector $x^{t}$ from the master. At this point they have their own update of $x_{vt}$, which we denote as $\bar{x}_{vt}$. The $v$-th worker then combines $\bar{x}_{vt}$ and $x^{t}$ to obtain $
x^{t+1}_v=\lambda_{vt} x^{t}+(1-\lambda_{vt})\bar{x}_{vt}$ where $0<\lambda_{vt}\leq 1$. Worker $v$ then uses $x^{t+1}_v$ to update its parameter vector and the process continues.   
Note that when $\lambda_{vt}=1$ (for all  $v,t$), the generalized version reduces to Anytime-Gradients since the computations conducted by  workers during idle periods are ignored. The $\lambda_{vt}$ should be chosen to speed convergence. Motivated by the convergence analysis in Sec. \ref{sec:varince}, we suggest using the following for $\lambda_{vt}$:
\begin{align}
\lambda_{vt} = \frac{\sum_{v=1}^N q_v}{\bar{q}_v+\sum_{v=1}^N q_v} 
\end{align}   
where $q_v$ is the number of iterations completed during the epoch by the $v$-th worker, and $\bar q_v$ is the number of iterations completed by worker $v$ during the worker-to-master-to-worker communication period. Note that each worker has to calculate $\lambda_{vt}$ in each epoch. The $\lambda_{vt}$ not necessary the same across epochs.  

We performed a numerical experiment using linear regression to evaluate the performance of generalized Anytime-Gradients. We used $10$ computing nodes in the Amazon EC2 cloud as workers and used a $500,\!000 \times 1000$ data matrix $A$. Each worker is given $50,\!000$ data vectors and we set $T=50$ secs. Fig. \ref{fig:fig17} plots the comparison of the normalized error of Anytime-Gradients with that of the generalized version. We observe that the generalized version converge to the correct solution faster than the original Anytime-Gradients approach. Faster convergence is due to addition iterations computed during the idle times. 
\section{Conclusion}
We proposed a parallelized SGD method named ``Anytime-Gradients". Although existing methods to parallelizing SGD work well in idealized settings, they fail to obtain the promised acceleration in many practical settings due to stragglers (slow working nodes). Anytime-Gradients exploits both stragglers and faster workers to realize a faster convergence. We provided a convergence analysis for our scheme. This is used to optimize the combining parameters used by the algorithm. We tested our scheme in Amazon EC2 cloud. Numerical results show significant improvement in comparison to previous methods. We are currently working to extend our analysis techniques to be able to characterize the performance of the generalized scheme  

\begin{figure}
	\centering
	\tikzset{every mark/.append style={scale=1.5}}
\begin{tikzpicture}[scale=0.7]
	\begin{semilogyaxis}[
		height=8cm,
		width=10cm,
		grid=major,
		xlabel=Epochs,
		ylabel=Normalized error,
		axis on top,xmin=1, xmax=9, ymin=0, ymax=1]
		\addlegendentry{Anytime-Gradients}
		\addplot [line width=0.5mm, color=blue, solid, every mark/.append style={solid, fill=blue},mark=diamond*] coordinates {
		(1,	1.00001)
		(2,	0.730602)
		(3,	0.533757)
		(4,	0.362715)
		(5,	0.258091)
		(6,	0.186433)
		(7,	0.136437)
		(8,	0.0999132)
		(9,	0.0734107)
		(10,0.052957)
		(11,0.0389812)
		
		};
		\addlegendentry{Generalized Anytime Gradient}
		\addplot [line width=0.5mm, color=black, dashed, every mark/.append style={solid, fill=black},mark=asterisk] coordinates {
		(1,	1.00001)
		(2,	0.331583)
		(3,	0.246619)
		(4,	0.184115)
		(5,	0.128228)
		(6,	0.0965429)
		(7,	0.0730043)
		(8,	0.0554984)
		(9,	0.0423459)
		(10,0.0325352)
		(11, 0.0249346)
		};
	\end{semilogyaxis}
\end{tikzpicture}
	\caption{Normalized error vs epoch for generalized Anytime-Gradients..}
	\label{fig:fig17}
\end{figure}
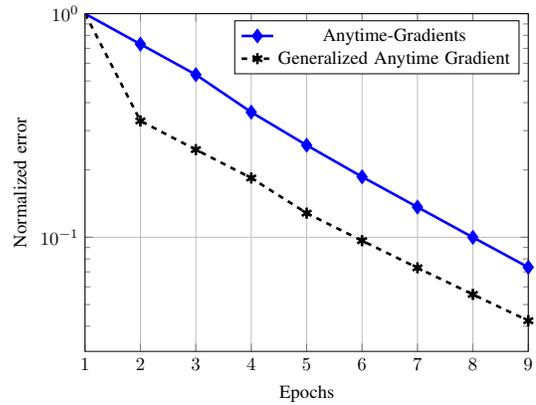

\bibliographystyle{IEEEtran}
\bibliography{reference} 
\appendices
\section{Proof of Theorem \ref{thm:expected.distance}}\label{apn.expected.distance}
We start by noting that
\begin{align}
F(x)  \leq \sum_{v=1}^{N}  \sum_{t=1}^{q_v} \frac{\lambda_v}{q_v}  F(x_{vt}).
\end{align}
This follows from the convexity of $F(x)$.  Recalling that $x^*$ is a global minimizer, we can write
\begin{align}
\label{eqn:main.therm}
F(x) -F(x^*) \leq \sum_{v=1}^{N} \frac{\lambda_v}{q_v} \sum_{t=1}^{q_v}   [F(x_{vt})-F(x^*)],
\end{align} 
since $\sum_{v=1}^{N}\lambda_v=1$ and $\lambda_v\geq 0$. The following theorem provides a bound on $\sum_{t=1}^{q_v}   [F(x_{vt})-F(x^*)]$.
{\theorem \label{thm.expected.distance.2} Let $\eta_{vt} = L+\beta_{vt}$ be the step size and let $s_{vt}=\nabla f(x_{vt},a_{{vt}})-\nabla F(x_{vt})$. Then
	\begin{align}
	\label{eqn:thm2}
	\sum_{t=0}^{q_v}& [F(x_{vt})-F(x^*)] \leq \!F(x_0) \!- \!F(x^*)\!+  \!(L\!+\beta_{vq_v}) \!D^2\nonumber \\ & +  \sum_{t=0}^{q_v-1}\frac{\|s_{vt}\|^2}{2 \beta_{vt}} + \sum_{t=0}^{q_v-1} \langle s_{vt},x^*-x_{vt} \rangle, \;\;\;\; \forall q_v 
	\end{align}}
\emph{Proof:} See Appendix \ref{apn.expected.distance.2}. 

Now we substitute \eqref{eqn:thm2} in \eqref{eqn:main.therm}: 
\begin{align}
\label{eqn:temp1}
&F(x) -F(x^*) \leq \sum_{v=1}^{N} \frac{\lambda_v}{q_v} \Bigg \{F(x_0) - F(x^*)+L D^2\nonumber   \\ &   + \beta_{vq_v} D^2+  \sum_{t=0}^{q_v-1}\left[\frac{\|s_{vt}\|^2}{2 \beta_{vt}} + \langle s_{vt},x^*-x_{vt} \rangle\right] \Bigg\}
\end{align} 
Recall that $s_{vt}=\nabla f(x_{vt},a_{vt})-\nabla F(x_{vt})$ and due to \eqref{eqn:unbiased}, we condition on the data used in the previous steps to get
\begin{align}
E_{a_{vt}}[\langle s_{vt},x^*-x_{vt}\rangle | a_{v0},\ldots a_{v{t-1}} ] = 0.
\end{align}
Finally we take the expectation of both sides of \eqref{eqn:temp1} to get
\begin{align}
&E[F(x) -F(x^*)] \nonumber \\ & \!\leq \!\sum_{v=1}^{N}\! \frac{\lambda_v}{q_v}\! \left \{\!F(x_0)\! - \!F(x^*) \!+\!(L\!+\!\beta_{q_v})\! D^2\! +\! \sum_{t=0}^{q_v\!-1}\! \frac{E\![\!\|s_{vt}\|^2]}{2\beta_{vt}} \! \right\}  \nonumber\\ & \!\leq \!\sum_{v=1}^{N} \!\frac{\lambda_v}{q_v}   \!\left \{\!F(x_0)\! -\! F(x^*)\!+\!(L\!+\!\beta_{vq_v})\! D^2\! +  \!\sum_{t=0}^{q_v\!-1} \!\frac{\sigma^2}{2\beta_{vt}}  \right\}.\nonumber
\end{align}  
With the substitution of $\beta_{vt}= \sqrt{t+1}\sigma/D$ and using the bound $\sum_{t=1}^{q_v}1/\sqrt{t} \leq 2\sqrt{q_v}-1$ we obtain Theorem \ref{thm:expected.distance}.

\section{Proof of Theorem \ref{thm.expected.distance.2}}
\label{apn.expected.distance.2}
The update rule in step 6 of Algorithm 2 is equivalent to solving following problem
\begin{align}
x_{vt} = \arg \min_{x \in X} \{ &\langle \nabla f_{v(t-1)}(x_{v(t-1)}, a_{v(t-1)}), x\rangle  \nonumber \\ &   +\eta_{v(t-1)} d(x,x_{v(t-1)} ) \}
\end{align}
where $ d(x, x_{v(t-1)} ) = 1/2\|x-x_{v(t-1)}\|^2_2$. In the remainder of this proof, we drop the subscript $v$ for notation convenience. We use the following lemma from \cite{Dekel:2012}

{\lemma Let $X$ be a closed convex set, $\phi(\cdot)$ a convex function on $X$, and let $d(x,u)=(1/2)\|x-u\|_2^2$ \; for $(x,u) \in X$. If
	\begin{align}
	x^+ =\arg \min_{x \in X} \{\phi(x) +d(x,u ) \}
	\end{align}   
	then
	\begin{align}
	\phi(x) + d(x,u) \geq \phi(x^+) + d(x^+,u) + d(x,x^+).
	\end{align}}
We are now ready to prove Theorem 4. We first define
\begin{align}
\label{eqn:lx}
l_t(x) = F(x_t) +\langle \nabla F(x_t), x-x_t\rangle, \;\;\;\; \forall t\geq 0
\end{align}
and 
\begin{align}
\label{eqn:hx}
h_t(x) &= F(x_t) + \langle \nabla f_{t}(x_t,a_{t}), x-x_t\rangle, \\ & = l_t(x) + \langle s_t,x-x_t\rangle,
\end{align}
where
\begin{align}
s_t=\nabla f(x_t,a_{t})-\nabla F(x_t).
\end{align}
Using the smoothness property of $F(x_t)$, \cite{Nesterov:2014}
\begin{align}
& F(x_t) \leq F(x_{t-1}) \nonumber \\& - \langle \nabla F(x_{t-1}), x_t-x_{t-1}\rangle   + \frac{L}{2} \|x_t-x_{t-1}\|^2\\ &
= h_{t-1}(x_t) \!-\!  \langle s_{t-1},x_t-x_{t-1}\rangle \!+ \frac{L}{2} \|x_t\!-\!x_{t-1}\|^2\\ &
= h_{t-1}(x_t) -  \langle s_{t-1},x_t-x_{t-1}\rangle \nonumber \\ &+ \frac{L+\beta_{t-1}}{2} \|x_t-x_{t-1}\|^2-\frac{\beta_{t-1}}{2} \|x_t-x_{t-1}\|^2\\ &
\leq h_{t-1}(x_t) + \|s_{t-1}\|\|x_t-x_{t-1}\|  \nonumber \\ & +\frac{L+\beta_{t-1}}{2} \|x_t-x_{t-1}\|^2-\frac{\beta_{t-1}}{2} \|x_t-x_{t-1}\|^2 \label{eq43}\\ &
= h_{t-1}(x_t) + \frac{L+\beta_{t-1}}{2} \|x_t-x_{t-1}\|^2 +\frac{\|s_{t-1}\|^2}{2\beta_{t-1}}  \nonumber \\ & -\left(\frac{\|s_{t-1}\|}{\sqrt{2\beta_{t-1}}} -\sqrt{\frac{\beta_{t-1}}{2}} \|x_t-x_{t-1}\|^2\right)^2  \\ &
\leq  h_{t-1}(x_t) + (L+\beta_{t-1}) d(x_t,x_{t-1}) + \frac{\|s_{t-1}\|^2}{2\beta_{t-1}}, \label{eqn:1bound}
\end{align}
where \eqref{eq43} is due to the fact that $ |\langle s_{t-1},x_t-x_{t-1}\rangle| \leq \|s_{t-1}\|\|x_t-x_{t-1}\|$ by the Cauchy-Schwarz inequality. Note that $\nabla h_{t-1}(x_t) = \nabla f_{t-1}(x_{t-1}, a_{t-1})$. Therefore, using Lemma 5 with $\phi(x) =h_{t-1}(x_t)$, and identifying $x=x^*, u=x_{t-1}, x^+=x_{t-1}$, we find that
\begin{align}
\label{eqn:2bound}
&h_{t-1}(x_{t}) + (L+\beta_{t-1}) d(x_t,x_{t-1}) \leq h_{t-1}(x^*) \nonumber \\ &+ (L+\beta_{t-1}) d(x^*,x_{t-1}) - (L+\beta_{t-1}) d(x^*,x_{t}).
\end{align}
Now, we combine \eqref{eqn:1bound} and \eqref{eqn:2bound} to get
\begin{align}
F(x_t)
&\leq  h_{t-1}(x^*) + (L+\beta_{t-1}) d(x^*,x_{t-1}) \nonumber \\ & - (L+\beta_{t-1}) d(x^*,x_{t}) + \frac{\|s_{t-1}\|^2}{2\beta_{t-1}}.
\end{align}
We substitute the bound \eqref{eqn:hx} to get
\begin{align}
F(x_t)&  
\nonumber \leq  l_{t-1}(x^*) + \langle s_{t-1},x^*-x_{t-1}\rangle+ \frac{\|s_{t-1}\|^2}{2\beta_{t-1}} \nonumber \\ &\!+ \!(L\!+\!\beta_{t-1}) d(x^*\!,\!x_{t-1})\! -\! (\!L\!+\!\beta_{t-1}\!) d(x^*,x_{t})  \\&
=l_{t-1}(x^*)  + (L+\beta_{t-1}) d(x^*,x_{t-1})  \nonumber \\ & - (L+\beta_{t}) d(x^*,x_{t}) + (\beta_t-\beta_{t-1}) d(x^*,x_{t})\nonumber \\ &+ \frac{\|s_{t-1}\|^2}{2\beta_{t-1}} + \langle s_{t-1},x^*-x_{t-1}\rangle  \\&
\leq \!l_{t\!-\!1}\!(x^*\!)  \!+ \!(\!L\!+\!\beta_{t-1}\!) d(\!x^*,x_{t-1}\!)   \!+(\!\beta_t\!-\beta_{t-1}\!)\! D^2\nonumber \\ &\!-\! (\!L\!+\!\beta_{t}\!) d(x^*\!,\!x_{t}\!)\!+\! \frac{\|s_{t\!-\!1}\|^2}{2\beta_{t\!-\!1}} \!+\! \langle\! s_{t\!-\!1},x^*\!-\!x_{t\!-\!1}\!\rangle \\&
\!\leq \!F\!(x^*\!)  \!+\! \!(L\!+\!\beta_{t\!-1}\!) d(x^*\!,\!x_{t-1}\!) \!+\! (\beta_t\!-\!\beta_{t-1}\!)\! D^2 \nonumber \\ & \!- \!(L\!\!+\!\!\beta_{t}\!) d(x^*\!,\!x_{t}\!)\!  + \!\frac{\|s_{t\!-\!1}\|^2}{2\beta_{t\!-\!1}} \!\!+\! \langle s_{t\!-\!1},x^*\!-\!x_{t\!-\!1}\!\rangle.
\end{align}
Summing from $t=1$ to $t=q_v$ we find that
\begin{align}
&\sum_{t=1}^{q_v} F(x_t) 
\nonumber \\ & \!\leq\! q_v F(x^*)  \!+\! (L\!+\!\beta_{0}) d(x^*\!,\!x^{0}) \! -\!(L\!+\!\beta_{q_v}) d(x^*\!,\!x^{q_v}) \nonumber \\ &+ (\beta_{q_v}-\beta_{0}) D^2 + \sum_{t=0}^{q_v-1} \left[\frac{\|s_{t}\|^2}{2\beta_t} + \langle s_{t},x^{*}-x_{t}\rangle \right] \\ &
\leq q_v F(x^*)  + (L+\beta_{0}) d(x^*,x^{0}) + (\beta_{q_v}-\beta_{0}) D^2 \nonumber \\ &+\sum_{t=0}^{q_v-1} \left[ \frac{\|s_{t}\|^2}{2\beta_t} + \langle s_{t},x^{*}-x_{t}\rangle \right]. 
\end{align}
By noting that $d(x^*,x^{0}) \leq D^2$ and adding $F(x_0)-F(x^*)$ to both sides, we arrive at
\begin{align}
\label{eqn:themre4}
&\sum_{t=0}^{q_v} [F(x_t) \!-\! F(x^*)] \leq F(x_0)\! -\! F(x^*) \!+ \! (L\!+\!\beta_{q_v}) D^2 \nonumber \\&+ \sum_{t=0}^{q_v-1} \left[ \frac{\|s_{t}\|^2}{2\beta_t} + \langle s_{t},x^{*}-x_{t}\rangle \right].
\end{align}
which completes the proof of Theorem \ref{thm.expected.distance.2}.
\section{Proof of Theorem \ref{thm.variance}}
\label{apn:variance}
Assume that
\begin{align}
E[\|s_{vt}\|^2] \!=\! E[\|\nabla f(x_{vt}\!,\!a_{vt})\!-\!\nabla F(x_{vt})\|^2] \!=\! \sigma_{vt}^2 \!\leq\! \sigma^2.
\end{align}
Assume that $\|\nabla f(x,a)\|\leq G$ for all $x \in X$ so that $\|s_{vt}\|^2 \leq 4G^2$.  By the Cauchy-Schwarz inequality,
\begin{align}
\label{eqn:bound.s}
|\langle s_{vt}, x^*-x_{vt} \rangle | \leq \|s_{vt}\|\|x^*-x_{vt}\| \leq 4DG.
\end{align}

Let $\mathcal V[\cdot]$ denote the variance of the argument. We rewrite \eqref{eqn:temp1} as
\begin{align}
\label{eqn:var0}
&F(x) -F(x^*)\nonumber \\ & \leq \mathcal \sum_{v=1}^{N} \frac{\lambda_v}{q_v} \Bigg \{F(x_0) - F(x^*)+ (L+\beta_{q_v}) D^2 \nonumber \\ & \!+\!  \sum_{t=0}^{q_v-1} \!\left[ \frac{\|s_{vt}\|^2\!-\!\sigma_{vt}^2}{2\beta_{vt}} \!+\! \langle s_{vt}\!,\!x^*\!-\!x_{vt}\rangle \!+\!\frac{\sigma_{vt}^2}{2\beta_{vt}}\right] \Bigg\}.
\end{align}
Computing the variance of \eqref{eqn:var0} we find that
\begin{align}
\label{eqn:var1}
& \mathcal V[F(x) -F(x^*)] \nonumber \\ & \leq \mathcal V\Bigg[\sum_{v=1}^{N} \frac{\lambda_v}{q_v} \Bigg \{F(x_0) - F(x^*)+ (L+\beta_{q_v}) D^2 \nonumber \\ &+  \sum_{t=0}^{q_v-1} \left[ \frac{\|s_{vt}\|^2-\sigma_{vt}^2}{2\beta_{vt}} + \langle s_{vt},x^*-x_{vt}\rangle +\frac{\sigma_{vt}^2}{2\beta_{vt}} \right] \Bigg\}\Bigg]\\&
= \sum_{v=1}^{N}\sum_{t=0}^{q_v-1} \frac{\lambda_v^2}{q_v^2}  \mathcal V\left[ \frac{\|s_{vt}\|^2-\sigma_{vt}^2}{2\beta_{vt}} + \langle s_{vt},x^*-x_{vt}\rangle \right]. \label{eqn:var2}
\end{align} 
Note that $E\left[ \frac{\|s_{vt}\|^2-\sigma_{vt}^2}{2\beta_{vt}} + \langle s_{vt},x^*-x_{vt}\rangle \right]=0$. Next consider the variance 
\begin{align}
&\mathcal V \left[ \frac{\|s_{vt}\|^2-\sigma_{vt}^2}{2\beta_{vt}} + \langle s_{vt},x^*-x_{vt}\rangle \right] \nonumber \\& \leq 
\frac{2 \mathcal V[\|s_{vt}\|^2]}{4\beta_{vt}^2} + 2 \mathcal V [\langle s_{vt},x^*-x_{vt}\rangle] \\ &  
\leq \frac{  E[\|s_{vt}\|^4]}{2\beta_{vt}^2} + 2  E [\langle s_{vt},x^*-x_{vt}\rangle^2] \\&
\leq \frac{  4G^2 E[\|s_{vt}\|^2]}{2\beta_{vt}^2} + 2 \|x^*-x_{vt}\|^2E[\|s_{vt}\|^2] \\&
\leq \frac{  2G^2 \sigma^2}{\beta_{vt}^2} + 4 D^2 \sigma^2. \label{eqn:partial.var}
\end{align}
Finally, we use $\beta_{vt}=\sqrt{t+1}\sigma/D$ and substitute \eqref{eqn:partial.var} in \eqref{eqn:var2} to find that
\begin{align}
\label{eqn:var3}
\mathcal V[F(x) -F(x^*)] &\leq \sum_{v=1}^{N}\frac{2\lambda_v^2\sigma^2D^2}{q_v^2} \sum_{t=1}^{q_v}   \left[\frac{  G^2 }{t \sigma^2} + 2\right]
\nonumber \\ & \leq \sum_{v=1}^{N}\frac{2\lambda_v^2\sigma^2D^2}{q_v^2} \sum_{t=1}^{q_v}   \left[\frac{  G^2 }{\sigma^2} + 2\right]
\\ & =\sum_{v=1}^{N}\frac{2\lambda_v^2\sigma^2D^2}{q_v}   \left[\frac{  G^2 }{\sigma^2} + 2\right].
\end{align}
This finishes the proof of Theorem \ref{thm.variance}. 

\section{Proof of Theorem \ref{thm.high.probability}}
\label{app.high.probability}
With the substitution of $\beta_{vt}= \sqrt{t+1}\sigma/D$ and using the bound $\sum_{t=1}^{q_v}1/\sqrt{t} \leq 2\sqrt{q_v}-1$, we rewrite \eqref{eqn:var0} as
\begin{align}
\label{eqn:var0a}
&F(x) -F(x^*)\nonumber \\ & \leq \sum_{v=1}^{N} \frac{\lambda_v}{q_v} \left \{F(x_0) - F(x^*)+LD^2 +  2 \sigma D \sqrt{q_v} \right \} \nonumber \\ & + \sum_{v=1}^{N}\sum_{t=0}^{q_v-1} \frac{\lambda_v}{q_v}\!\left[ \frac{\|s_{vt}\|^2\!-\!\sigma_{vt}^2}{2\beta_{vt}} \!+\! \langle s_{vt}\!,\!x^*\!-\!x_{vt}\rangle\right].
\end{align}
Let $Y$ be the last part of \eqref{eqn:var0a}:
\begin{align}
Z  =  \sum_{v=1}^{N}  \sum_{t=0}^{q_v-1}\frac{\lambda_v}{q_v} \!\left[ \frac{\|s_{vt}\|^2\!-\!\sigma_{vt}^2}{2\beta_{vt}} \!+\! \langle s_{vt}\!,\!x^*\!-\!x_{vt}\rangle \!\right].
\end{align}
From \eqref{eqn:var3}, the variance bound of $Z$ is
\begin{align}
\label{eqn:var3a}
\mathcal V[Z] \leq\sum_{v=1}^{N}\frac{2\lambda_v^2\sigma^2D^2}{q_v}   \left[\frac{  G^2 }{\sigma^2} + 2\right].
\end{align} 
Let 
\begin{align}
z_{vt}=\frac{\lambda_v}{q_v}\left(\frac{\|s_{vt}\|^2\!-\!\sigma_{vt}^2}{2\beta_{vt}} \!+\! \langle s_{vt}\!,\!x^*\!-\!x_{vt}\rangle \!+\!\frac{\sigma_{vt}^2}{2\beta_{vt}}\right)
\end{align}
so that $Z=\sum_{v=1}^{N} \sum_{t=0}^{q_v-1} z_{vt}$. Let $\gamma=\max_v{\frac{\lambda_v}{q_v}}$. Based on our assumption $\|s_{vt}\|^2 \leq 4G^2$ and \eqref{eqn:bound.s}, we bound $|z_{vt}|$:
\begin{align}
|z_{vt}| \leq \gamma 2GD \left(\frac{G}{\sigma}+2\right), \;\; \forall v, t.
\end{align}

Note that $z_{vt}$ is a random variable of data samples $a_{v0},a_{v1},\ldots, a_{v(t-1)}$. We can show that
\begin{align}
E_{a_{vt}}[z_{vt}|a_{v0},a_{v1},\ldots, a_{v(t-1)}] = 0.
\end{align}
As workers sample independently, it can be easily shown
\begin{align}
E_{a_{vt}}[z_{vt}|a_{u\bar t}, \forall u \in\{1,\ldots v-1\} , \bar t \in \{0,\ldots t-1\}] = 0.
\end{align}
Therefore $z_{10}, \!z_{11}, \!\ldots,\! z_{1q_1},\! z_{20},\! \ldots\! z_{2q_2},\! \ldots,\! z_{N0},\! \ldots, \!z_{Nq_N}$ forms a martingale difference sequence with respect to $a_{10}, a_{11}, \ldots, a_{1q_1}, a_{20}, \ldots a_{2q_2}, \ldots, a_{N0}, \ldots, a_{Nq_N}$. Thus for any $\delta\in(0,1)$, based on the Lemma A.8 of \cite{Cesa-Bianchi:2006} and \cite{Dekel:2012}, the following holds with probability at least $1-\delta$:
\begin{align}
\label{eqn:Z}
Z \leq \gamma 2GD \left(\frac{G}{\sigma}+2\right) \log(1/\delta) \sqrt{1+\frac{18 \mathcal V[Z]}{\log(1/\delta)}}.
\end{align}   
We substitute \eqref{eqn:Z} in \eqref{eqn:var0} to get
\begin{align}
\label{eqn:var01}
&F(x) -F(x^*)\nonumber \\ & \leq \sum_{v=1}^{N} \frac{\lambda_v}{q_v} \left \{F(x_0) - F(x^*)+LD^2 +  2 \sigma D \sqrt{q_v} \right \} \nonumber \\ & + \gamma 2GD \left(\frac{G}{\sigma}+2\right) \log(1/\delta) \sqrt{1+\frac{18 \mathcal V[Z]}{\log(1/\delta)}}.
\end{align}
To prove Theorem \ref{thm.high.probability}, we subtract  $E[F(x) -F(x^*)]$ (given in Theorem \ref{thm:expected.distance}) from \eqref{eqn:var01} and use \eqref{eqn:var3}. This completes the proof of Theorem \ref{thm.high.probability}.


\end{document}